\documentclass{article}
\usepackage{ijcai19}

\usepackage{times}
\usepackage{soul}
\usepackage{url}
\usepackage[hidelinks]{hyperref}
\usepackage[utf8]{inputenc}
\usepackage[small]{caption}
\usepackage{graphicx}
\usepackage{multirow}
\usepackage{amsmath}
\usepackage{amsfonts}
\usepackage{booktabs}
\usepackage{algorithm}
\usepackage{algorithmic}
\newcommand{\model}{\text{FATHOM}}

\title{Federated Multi-task Hierarchical Attention Model for Sensor Analytics}

 \author{
 Yujing Chen$^1$
 \and
 Yue Ning$^2$\and
 Zheng Chai$^1$\and
 Huzefa Rangwala$^1$
 \affiliations
 $^1$Department of Computer Science,George Mason University,Virginia, USA\\
 $^2$Department of Computer Science, Stevens Institute of Technology,New Jersey, USA\\
 \emails
 $^1$\{ychen37,zchai2,hrangwal\}@gmu.edu\\
 $^2$\{yue.ning\}@stevens.edu
 }
\begin{document}

\maketitle

\begin{abstract}
Sensors are an integral part of modern Internet of Things (IoT)  applications. 
%
There is a critical need for the analysis of heterogeneous multivariate temporal data obtained from the individual sensors of these systems.
In this paper we particularly focus on the problem of the scarce amount of training data 
available per sensor. We propose a novel \textbf{f}eder\textbf{a}ted multi-\textbf{t}ask \textbf{h}ierarchical attenti\textbf{o}n \textbf{m}odel (\model) that jointly 
trains classification/regression models from multiple sensors. 
The attention mechanism of the proposed model seeks to extract feature representations from the input and learn a shared representation focused on time dimensions across multiple sensors. 
The underlying temporal and non-linear 
relationships are modeled using a combination of attention mechanism and long-short term memory (LSTM) networks. 
We find that our proposed method outperforms a wide range of competitive baselines in both classification and regression settings on 
activity recognition and environment monitoring datasets. 
We further provide visualization of feature representations learned by our model at the input sensor level and central time level. 
\end{abstract}

\section{Introduction}

Ubiquitous sensors seek to improve 
the quality of everyday
life
through pervasively interconnected objects. 
%
As an example, consumer-centric healthcare devices provide accurate and personalized feedback on individuals' health conditions.
A wide array of  sensors in the form of wearable
devices (e.g., clothing and wrist worn devices), smartphones, and infrastructure components (e.g., low-cost
sensors, cameras, WiFi and work stations) are the chief enablers. These solutions commonly referred to as the
Internet of Things (IoT) allow for fine-grained sensing and inference of users' context, physiological
signals, and even mental health states. 
%
These sensing and detection capabilities coupled with advanced data analytics are leading to intelligent
intervention and persuasion techniques that provide an appealing end-to-end solution for 
various domains e.g., environment monitoring, healthcare, education, and workplace management.

%
%
%



Most data generated by these 
sensors are multivariate temporal sequences. 
We model the sequential data from sensors 
using recurrent neural networks (RNNs)~\cite{rumelhart1988learning} 
adapted for handling long-term dependencies; referred commonly 
as long short-term memory models (LSTM)~\cite{hochreiter1997long}.
%
%
%
Further individual sensors usually do not have enough training data for learning generalizable models. 
In this study we aim to address this particular challenge within the 
framework of federated multi-task learning (MTL). 
This allows for multiple benefits: (i) reduced communication costs by not sending 
large volumes of data to a central server, (ii) distributed learning and (iii) privacy 
guarantees when the data resides only on the client. 
Similar to prior work on federated multi-task learning \cite{smith2017federated},  data associated with 
each task stays local. Individual models are learned for each task first, and a central node controls 
the communication between local model parameters and global shared 
parameters.  
We present a novel federated multi-task approach that uses a hierarchical attention mechanism to 
learn a common representation among multiple tasks.  
%
The inter-feature correlations of each individual task are captured by 
sensor-specific attention layers applied on input sensors' time series. Meanwhile, the temporal correlations are 
captured by attending to all tasks across the time dimension.
%
%
%
%
The scope of this paper involves modeling the input from multiple heterogeneous sensors across multiple users. The proposed 
algorithm is federated in the sense that no data leaves the user. However, we assume that data from 
each of the sensors are available at the same time.

%
%
The key contributions of this paper can be summarized as follows: 
\begin{itemize}
    \item We propose a novel federated multi-task hierarchical attention model (\model) that extracts feature representations with multiple attention aspects. This leads to 
    improved classification and regression performance for input temporal data from multiple sensors. 
    \item Our proposed approach outperforms a wide range of baselines on two multi-modal sensor datasets from different domains with multi-binary labels or multi-continuous labels.
    \item We present a framework to  extract and visualize key feature representations from sequential data.
\end{itemize}

\section{Related Work}

\subsection{Multi-task learning}

Multi-task learning (MTL) is
designed for the simultaneous training of
multiple related prediction tasks.  Leveraging common information 
across related tasks has shown to be effective in improving 
the generalization performance of each task~\cite{caruana1997multitask,evgeniou2005learning,bonilla2007kernel,pentina2017multi}.
MTL is particularly useful 
when there are a number of related tasks but some tasks have scarce amounts of
available training data. Many researchers have investigated MTL
from varied perspectives~\cite{chen2018meta,zhou2012multi}. 
Task relationships are modeled by 
sharing layers/units of neural networks~\cite{caruana1997multitask}, sharing a low dimensional subspace~\cite{argyriou2007multi}, or
assuming a clustering among tasks~\cite{zhou2011clustered,jacob2009clustered}.
In this work we do not make any assumptions on task relationships beforehand and learn the common representation with attention mechanisms directly from the data.

Specifically, for sensor analytics, a multi-task multilayer perceptron (MLP) model ~\cite{vaizman2018context} was developed to recognize
different human activities from mobile sensors. By manipulating the MLP model to fit uncontrolled in-the-wild 
unbalanced data, this approach outperformed a standard logistic regression (LR) model~\cite{vaizman2017recognizing}. The MLP and  LR approaches 
do not leverage the underlying temporal dependencies and inter-feature correlations of sensors' data.
  



\subsection{Federated learning}
The target of federated learning is to train a centralized model while training data 
is distributed on separate client 
nodes. Prior research on federated learning aims at learning a single model 
across the network~\cite{mcmahan2016communication,konevcny2016federated,konevcny2015federated}. Different from these 
works, Smith \textit{et al.}~\cite{smith2017federated} provided an approach to solve statistical challenges in the 
federated setting
	of multi-task learning. The key objectives are solving issues of high
	communication costs, stragglers (nodes with less computation power), and reliability. We adopt federated multi-task learning using a hierarchical attention mechanism. 
%
In this work, we focus on improving model performance across all local tasks. 


\subsection{Attention-based deep network}
Fundamentally, neural networks allocate 
importance to input features through the weights of the learned model. 
In the context of deep learning, an attention 
mechanism allows a network 
to assign different levels of 
importance 
to various inputs by adjusting the weights. This 
leads to a better feature representation. 
%
%
Attention approaches 
can be roughly divided into Global Attention and Local Attention~\cite{luong2015effective}. The global attention is akin  
to soft attention~\cite{bahdanau2014neural,xu2015show,yao2015video}; where
the alignment weights are learned and placed ``softly'' over all patches in 
the source data. Local attention only selects one patch of the data to access at a time.  



Multi-level attentions are studied for improving  
document classification~\cite{yang2016hierarchical} and predicting spatio-temporal 
data~\cite{liang2018geoman}. 

Our proposed hierarchical attention approach learns individual feature correlations within each local task. It also learns shared feature representations
	across distributed tasks with a central attention mechanism that focuses on time steps. By passing this central time attention back to each local task, we are able to learn 
    task-specific representations.

 

\section{Methods}

\subsection{Problem Definition and Notations}

Suppose we are given $K$ tasks with their input data $\{(\mathbf{X}^{(1)}, \mathbf{Y}^{(1)}), (\mathbf{X}^{(2)}, \mathbf{Y}^{(2)}),..., (\mathbf{X}^{(K)}, \mathbf{Y}^{(K)})\}$, generated by $K$ distributed nodes. Each task 
 $k$ consists of instances collected from multiple sensors and each sensor provides multiple features. Assuming we have $D$ features in total from all the sensors, $\mathbf{X}_{i}^{(k)} \in \mathbb{R}^D$ represents the features for the $i^{th}$ instance, 
 $\mathbf{Y}_{i}^{(k)} \in \mathbb{R}^{M}$ is the corresponding label vector, $M$ is the number of labels in task $k$. 
 With labeling function $\mathcal{F}$ $:\mathbf{X} \rightarrow \mathbf{Y}$, the goal is to 
 learn a model from
 a hypothesis set $\mathcal{S} \subset \{s:\mathbf{X} \rightarrow \mathbf{Y} \}$ using 
 the training data that minimizes 
 the average error across all the $K$ tasks: 
\setlength{\belowdisplayskip}{0pt} \setlength{\belowdisplayshortskip}{0pt}
\setlength{\abovedisplayskip}{0pt} \setlength{\abovedisplayshortskip}{0pt}
    \begin{equation} \label{1.1}
 	\mathbf E_r = \frac{1}{K} \sum\limits_{k=1}^K \mathcal{L}(s_k(\mathbf{X}), \mathcal{F}_k(\mathbf{X}))
	\end{equation}

For multi-label classification problems, we train a network to minimize the cross-entropy of the predicted and true distributions for each task with $N$ instances and each instance has $M$ labels as:
\setlength{\belowdisplayskip}{0pt} \setlength{\belowdisplayshortskip}{0pt}
\setlength{\abovedisplayskip}{0pt} \setlength{\abovedisplayshortskip}{0pt}
	\begin{equation} \label{1.3}  
 	\mathbf \ \\  \mathcal{L}(\hat{\mathbf{Y}}, \mathbf{Y}) = -\sum\limits_{m=1}^M \sum\limits_{i=1}^N \hat{y}_{i}^m log(y_{i}^m)
	\end{equation}

where $\mathcal{L}(\hat{\mathbf{Y}}, \mathbf{Y})$ is the loss function, $\hat{\mathbf{Y}}$ denotes the predicted labels, and 
$\mathbf{Y}$ is the true labels. For multi-output regression tasks, we train the networks to minimize the mean absolute error of the predicted and true distributions for each task with $N$ instances and each instance has $M$ outputs as: 
	\begin{equation} \label{1.3} 
 	\mathbf \ \mathcal{L}(\hat{\mathbf{Y}}, \mathbf{Y}) = \frac{1}{M}\sum\limits_{m=1}^M \sum\limits_{i=1}^N |\hat{y}_{i}^m-y_{i}^m| 
	\end{equation}

\subsection{Proposed Model}

\begin{table}
\centering
\caption{Notations. Lower case letters represent scalars, bold upper case letters represent matrices, bold lower case letters represent vectors. }
\label{tab:booktabs}
\begin{tabular}{l|l}  
\toprule
Notation  & Meaning\\
\midrule
  K & \# of tasks \\
  D & \# of features in each task \\
  M & \# of labels in each task \\
  $\mathbf{X}^{(k)}$ & task $k$\\
  $\mathbf{X}_{T}^{(k)}$ & task $k$ with window size $T$\\
  $\mathbf{a}_{d}^{(k)}$, $a_t$ & attention weight \\
  $\mathbf{c}_{d}^{(k)}$ & sensor-specific level context vector\\
$\mathbf{c}_{t}^{(k)}$ & central time context vector\\
\bottomrule
\end{tabular}

\end{table}

Our proposed model includes two main components: 1) Sensor-specific attention is applied directly on the input data of each task to learn the importance of each feature.
This is followed by a LSTM layer to learn the captured feature representations. The last LSTM layer is designed to capture the learned central feature representations followed by two fully connected layers to predict the task labels. 2) A central time attention is applied on time steps of the concatenated tasks to capture importance of time across all tasks. Our method is detailed in Algorithm~\ref{alg:algorithm}.

\begin{figure}[h]
 	\begin{center}
	\includegraphics[trim={1.8cm 0cm 8cm 0cm},clip, width = 9cm]{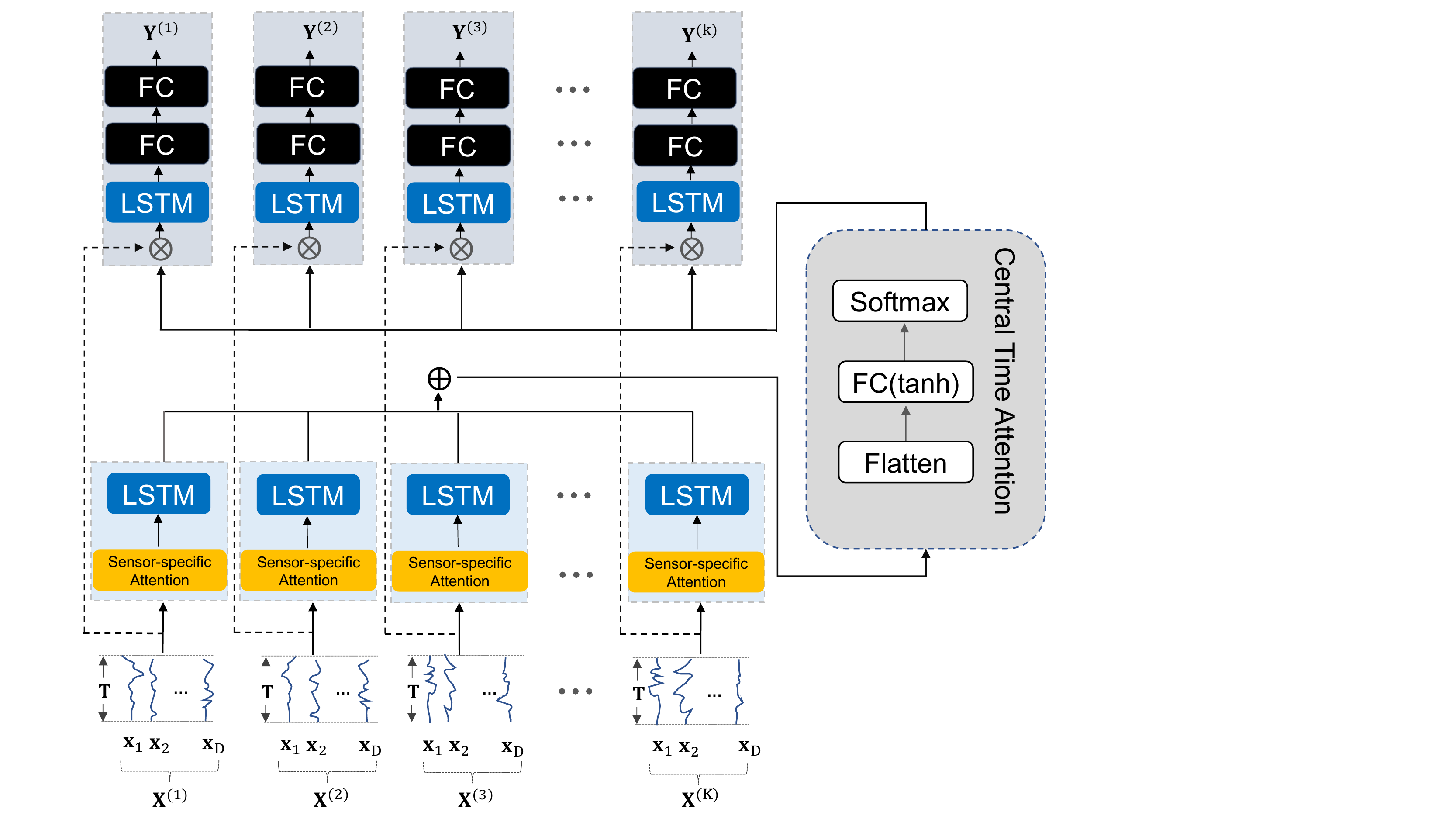}
	\caption{Architecture illustration of proposed federated Multi-task Hierarchical Attention Model (\model). `FC' indicates fully connected layer, $\oplus$ indicates concatenation and $\otimes$ indicates element-wise multiplication. }\label{fig:framework}
	\end{center}
	\end{figure}

In each task, the input series has $D$ features, $\mathbf{X}^{(k)} = \{\mathbf{x}_{1},\mathbf{x}_{2},\mathbf{x}_{3}, ..., \mathbf{x}_{D} \} \in \mathbb{R}^{N \times D}$. 
The sensor-specific attention at each task node obtains the attention weight of each series with time-step window length $T$ with a softmax function. It then gets the attention distributions on feature dimension by multiplying the raw features of each time-step window and the attention weight. The attention vector is obtained by passing the attention distributions to one fully connected layer with a $\tanh$ activation. The obtained attention vector is fed into a LSTM layer to further capture the learned feature representation. 
The central time attention aims to extract a shared representation across all input tasks in the time window. By concatenating all hidden states after LSTM layers of each task, we apply a flatten function, an activation by $\tanh$, and a softmax function to get the shared attention weights. By multiplying raw features at each time-step window of each task and the shared attention weights, we get the attention score of each task. A second LSTM layer is incorporated to learn the second pass after central time attention is applied on each task. Finally a classifier is used to predict the labels of each task. An illustration of our model structure can be found in Figure ~\ref{fig:framework}.

In the following sections we will describe each part of our model in detail.

\begin{algorithm}[tb]
\algsetup{linenosize=\small}
\caption{Learning for \model}
\label{alg:algorithm}
\textbf{Input}: data of $K$ tasks\\
\begin{algorithmic}[1] 
\FOR{$k \in \{1,2,...,K\}$ in parallel over $K$ nodes}
	\STATE calculate attention vector $\mathbf{c}_{T(1)}^{(k)}$
	\STATE pass $\mathbf{c}_{T(1)}^{(k)}$ to a LSTM layer to get $\mathbf{h}^k$
\ENDFOR
\STATE calculate $\mathbf{s}_T$ $\gets$ $\mathbf{h}^{1} \oplus \mathbf{h}^{2} ... \oplus \mathbf{h}^{K}$
\STATE compute central attention $\mathbf{a}_{1\rightarrow T}$ based on $\mathbf{s}_T$
\FOR{$k \in \{1,2,...,K\}$ in parallel over $K$ nodes}
	\STATE update attention vector $\mathbf{c}_{T(2)}^{(k)}$ $\gets$ $\mathbf{X}^{(k)}*\mathbf{A}$ 
\ENDFOR
\STATE \textbf{return} $\mathbf{Y}^{(1)}, \mathbf{Y}^{(2)},..., \mathbf{Y}^{(K)}$ $K$ label matrices
\end{algorithmic}
\end{algorithm}

\subsubsection{Sensor-specific Attention}

Feature extraction at the input sensor level can increase the probability of capturing features that are related to  target labels. These features should be given higher weights than other features while computing a sensor-specific representation. It is hard to tell which part of the features has predictive information, so we choose a global attention mechanism~\cite{luong2015effective} to capture feature representations by attending to all input features from each task. 

	
Assume a given task $k$ with $D$ input series as $\mathbf{X}^{(k)} = \{\mathbf{x}_{1},\mathbf{x}_{2},\mathbf{x}_{3}, ..., \mathbf{x}_{D} \} \in \mathbb{R}^{T \times D}$, where $T$ is the time-step window size, and $\mathbf{X}_{;,d}^{(k)} = [{x}_{1d},{x}_{2d},{x}_{3d}, ..., {x}_{Td} ]^\top \in \mathbb{R}^{T} $ is the $d$-th series with time-step window size $T$.
First we transform each series of $\mathbf{X}_{;,d}^{(k)}$ using a fully connected layer with $D$ units to obtain a hidden representation $ \mathbf{w}_{d}^{(k)}$.

	\begin{equation} \label{1.3} 
 		 \mathbf{w}_{d}^{(k)} = \mathbf{W}^{(k)} \mathbf{X}_{;,d}^{(k)}
	\end{equation}
 
In which, $\mathbf{W}^{(k)}$ is the trainable weight matrix. We compute the attention weight of the $d$-th input feature by applying a softmax function.  

	\begin{equation} \label{1.3} 
 		\textbf \\ \mathbf{a}_{d}^{(k)} = \frac{\exp( \mathbf{w}_{d}^{(k)} )}{\sum\limits_{d=1}^{D}  \exp(\mathbf{w}_{d}^{(k)})} 
	\end{equation}

We measure the importance of features by computing the context vector with the element-wise multiplication of $\mathbf{X}_{;,d}^{(k)}$ and the attention weights $\mathbf{a}_{d}^{(k)}$. Then a $\tanh$ activation is applied to get the final attention vector.

	\begin{equation} \label{1.3} 
 		 \mathbf{c}_{d}^{(k)} = \tanh ( \mathbf{X}_{;,d}^{(k)} \otimes \mathbf{a}_{d}^{(k)} )
	\end{equation}
	
\subsubsection{Central Time Attention} 

The attention distribution captured at sensor-specific levels focus on a specific part of features, which can only reflect the label information at a current timestamp. However, for time series data, there is tight temporal correlation of instances. It is essential to capture the hidden information of time. The central attention component aims at learning a shared representation across all tasks at each time step. Let $\mathbf{h}^{k}$ be the hidden representation of task $k$ after the first LSTM layer. First we concatenate the hidden representation across $K$ tasks to get the shared hidden representation $\mathbf{s}_{T}$:

    \begin{equation} \label{1.3} 
 		 \mathbf{s}_T = \mathbf{h}^{1} \oplus \mathbf{h}^{2} \oplus \mathbf{h}^{3} ... \oplus \mathbf{h}^{K}
	\end{equation}

We pass the shared hidden representation to a flatten layer to get a flattened hidden representation $\mathbf{f}$. 
    
    \begin{equation} \label{1.3} 
 		 \mathbf{f} = \text{flatten}(\mathbf{s}_T)
	\end{equation}
	
	Different from input-level attention, we apply a $\tanh$ nonlinearlity before softmax. By transforming $\mathbf{f}$ using a fully connected layer with $T$ units, $\tanh$ is applied to obtain the time-step level context vextor $\mathbf{u} \in \mathbb{R}^T$: 
	


	\begin{equation} \label{1.3} 
 		 \mathbf{u}= \tanh(\mathbf{f})
	\end{equation}

Then we compute the attention vector using a softmax function for each time stamp $t=1,..,T$: 

	\begin{equation} \label{1.3} 
 		\mathbf\\ a_t = \frac{\exp({u}_{t \in T})}{\sum\limits_{t \in T}  \exp({u}_{t})} 
	\end{equation}

The attention score $a_{t}$ is obtained by normalizing the context score ${u}_{t}$ at each time step $t$. We then repeat the attention weight $D$ times to get the attention matrix across input dimension. With permutation we obtain the attention vector $\mathbf{a}_t\in \mathbb{R}^{D}$. 

We measure the importance of each time-step by computing the attention matrix with an element-wise multiplication of $\mathbf{X}^{(k)}_t$ and the attention vector $\mathbf{a}_t$.	


	\begin{equation} \label{1.3} 
 		 \mathbf{c}_{t}^{(k)} = \mathbf{X}^{(k)}_t \otimes  {\mathbf{a}_t}
	\end{equation}

Here we obtain the extracted hidden representation across all tasks at the time-step level. By feeding $\mathbf{c}_{t(2)}^{(k)}$ of task $k$ to a LSTM layer and two fully connected layers, we get the predicted labels of each task.
	
\section{Experiments}
\subsection{Datasets}
To evaluate the performance of our approach we use several real-world datasets
 that have previously been used 
 in multi-task learning frameworks for sensor analytics. 

\begin{itemize}
    \item \textbf{ExtraSensory Dataset \footnote{http://extrasensory.ucsd.edu/}:}  Mobile phone sensor data (e.g., high-frequency motion-reactive sensors, location services, audio, watch compass) and watch sensor data (accelerator) collected from 60 users; performing any of 51 activities~\cite{vaizman2018extrasensory}. We select 40 users with at least 3000 samples  
    and use the provided 225-length feature vectors of time and frequency domain variables generated for each instance. We model each user as a separate task and predict their activities (e.g.,walking, taking, running). 
    \item \textbf{Air Quality Data \footnote{https://biendata.com/competition/kdd\_2018/data/}:} Weather data collected from multiple weather sensors (e.g., thermometer, barometer) from
    9 areas in Beijing 
    from Jan 2017 to Jan 2018. We model each area as a separate task; and 
    use the Observed Weather Data to predict the measure of 
    air pollutants (e.g., PM2.5, PM10) from May 1st, 2018 to May 31st, 2018. 
    
 \end{itemize}

\subsection{Comparative Methods}

We compared the proposed \model approach to several 
single-task learning and multi-task learning approaches. In particular, we 
consider the Logistic Regression (LR) model in~\cite{vaizman2017recognizing} as the single-task learning baseline.  We also 
use the multi-layer perceptron 
MLP (16,16) multi-task model in~\cite{vaizman2018context}. 
 
Convolutional Neural Networks is able to 
extract short-term basic patterns in time dimension
and find 
local dependencies among features.
This is referred by 
Convolutional Recurrent Neural Network (CRNN)~\cite{cirstea2018correlated} where we feed each input task to a separate CNN layer and replace the RNN layer with LSTM layer.
  
Besides our proposed \model~approach, we provide several other models for comparison.
\begin{itemize}
\item We use a single LSTM layer for single-task prediction as a comparison with LR approach. We refer to this by  S-LSTM.
\item We replace MLP(16,16) with one layer LSTM which can better capture long-term dependencies in learning. We refer to this model as 
Multi-task LSTM model (M-LSTM). 
\item We also perform an ablation study to assess the strengths of the different attention layers introduced in our proposed \model.  \model-sa is 
 without the Sensor-specific Attention and \model-ca is  without the Central Time Attention.
\end{itemize}

We compare the performance of these models with the proposed \model~model. For the 
classification datasets, the labels are highly unbalanced and hence we report the F1 score, precision, recall, and 
Balance Accuracy (BA) from~\cite{vaizman2018context}. For the regression dataset
we evaluate the performance using Symmetric mean absolute percentage error (SMAPE). For each experiment, we 
split our dataset into training data, validation data and test data, in the proportions of 60\%, 20\% and 20\%, respectively.

\subsection{Hyper-parameters}

Based on the performance on the validation set we choose the best group
of parameters, retrain a model with the identified parameters, and report results on the test set.  We set 
the hidden units of the both LSTM layers to 64, and both the regular dropout and the recurrent 
dropout to  0.25. We also impose $l2$ constraints on the weights within LSTM nodes to further reduce overfitting. We use a 
batch size of 60 for training and an initial learning rate of 0.001. We employ early stopping with patience value of 
$20$. For classification tasks,  we use categorical cross entropy to monitor the loss within the 
Adam optimization algorithm. For regression tasks, we change the loss to Mean absolute error (MAE).

\begin{table}
\centering
\scriptsize
\caption{Comparative Performance on the  ExtraSensory Dataset (Classification) and Air Quality Dataset (Regression). M-LSTM and S-LSTM use 
one layer LSTM for multi-task learning and single-task learning, respectively. Pr and Rec denote Precision and Recall, respectively.}\label{tab:com-result}
\begin{tabular}{l|llll|c}
\hline
\multirow{2}{*}{Methods} & \multicolumn{4}{c|}{ExtraSensory}   & \multicolumn{1}{l}{Air Quality} \\ \cline{2-6} 
    & Pr & Rec& F1 & BA  & SMAPE ($\downarrow$)                                                        \\ \hline
LR\cite{vaizman2017recognizing}                    & 0.57          & 0.60          & 0.52          & 0.72          & 1.23                                                           \\
MLP(16,16)\cite{vaizman2018context}            & 0.55          & 0.61          & 0.58          & 0.76          & 0.65                                                           \\
S-LSTM               & 0.79          & 0.71          & 0.74          & 0.84          & 0.52                                                           \\
M-LSTM                 & 0.45          & 0.62          & 0.52          & 0.77          & 0.66                                                           \\
CRNN\cite{cirstea2018correlated}                    & 0.43          & 0.68          & 0.54          & 0.78          & 0.64                                                           \\
\model (Proposed)                    & \textbf{0.89} & \textbf{0.77} & \textbf{0.82} & \textbf{0.88} & \textbf{0.45}                                                  \\ \hline
\end{tabular}
\end{table}

\begin{table}
\centering
\small
\caption{Ablation Study Showing Performance of \model \ Variants with Different Attention Aspects. \model-sa is without the Sensor-specific Attention and \model-ca is without the Central Time Attention. Pr and Rec denote Precision and Recall, respectively.}
\label{tab:com-attentions}
\begin{tabular}{l|llll|c}
\hline
\multirow{2}{*}{Methods} & \multicolumn{4}{c|}{ExtraSensory}  & \multicolumn{1}{l}{Air Quality} \\ \cline{2-6} 
                         & Pr     & Rec        & F1            & BA            & SMAPE ($\downarrow$)                                  \\ \hline
\model-ca                 & 0.50          & 0.61          & 0.54          & 0.77          & 0.73                                    \\
\model-sa                 & 0.80          & 0.69          & 0.74          & 0.84          & 0.51                                    \\
\model                    & \textbf{0.89} & \textbf{0.77} & \textbf{0.82} & \textbf{0.88} & \textbf{0.45}                           \\ \hline
\end{tabular}
\end{table}

%
%

\section{Results}
\subsection{Comparative Performance}

We demonstrate the prediction 
performance of 
the proposed \model~approach in comparison to different 
baseline approaches. 
Tables~\ref{tab:com-result} 
shows the classification and regression performance for the different models on different benchmarks. We observe that  
\model~significantly outperforms
the other models in terms of precision, recall, F1, and Balance Accuracy metrics for classification tasks.
Given the highly unbalanced characteristic of the 
ExtraSensory Dataset, it is promising to observe 
that \model~shows a higher number of true positives and 
lower false positives. 
The proposed model outperforms all of the other
models in range of 0.08-0.30 on F1 score and
0.04-0.16 on  Balanced Accuracy. For the CRNN model, the 
CNN layer can extract distinctive features of the 
correlated multiple time series, which acts similar to the 
attention mechanism. We observe that CRNN performs slightly 
better than multi-task LSTM (M-LSTM) and LR approaches, but 
not better than other models. The feature correlations captured by 
CNN on each task is loose and cannot 
represent temporal dependency effectively.

%
%
From Table \ref{tab:com-result} results on Air Quality Data we also note that 
\model~outperforms all of the other models by lowering 
the Symmetric mean absolute percentage error in range 0.07-0.78.



\subsubsection{Attention vs. No attention}

To further 
study the function of the two attention mechanisms in our model we perform 
an ablation study 
by removing either the local sensor-specific attention layer
or the central time attention layer and 
denote them by 
\model-sa and \model-ca, respectively. 
Table \ref{tab:com-attentions} shows the prediction performance 
of these two variants of \model. We observe that  
\model-sa with the central time attention still 
achieves very good performance in comparison to other
baseline approaches. The \model-sa model  captures both 
the temporal correlation and a common feature
representation across all tasks. However, 
\model-ca  does not perform well 
and is similar to the CRNN approach. The feature representations
learned by sensor-specific layer fail to leverage the temporal 
correlations in the input data. 
The  main \model~model with both attention mechanism outperforms 
\model-ca and \model-sa by 0.28 and 0.08 with respect to the F1-score on 
the Extrasensory dataset, respectively. 


	\begin{figure}[h]
 	\begin{center}
	\includegraphics[trim={2.5cm 5cm 11.5cm 4cm},clip, width = 10.5cm]{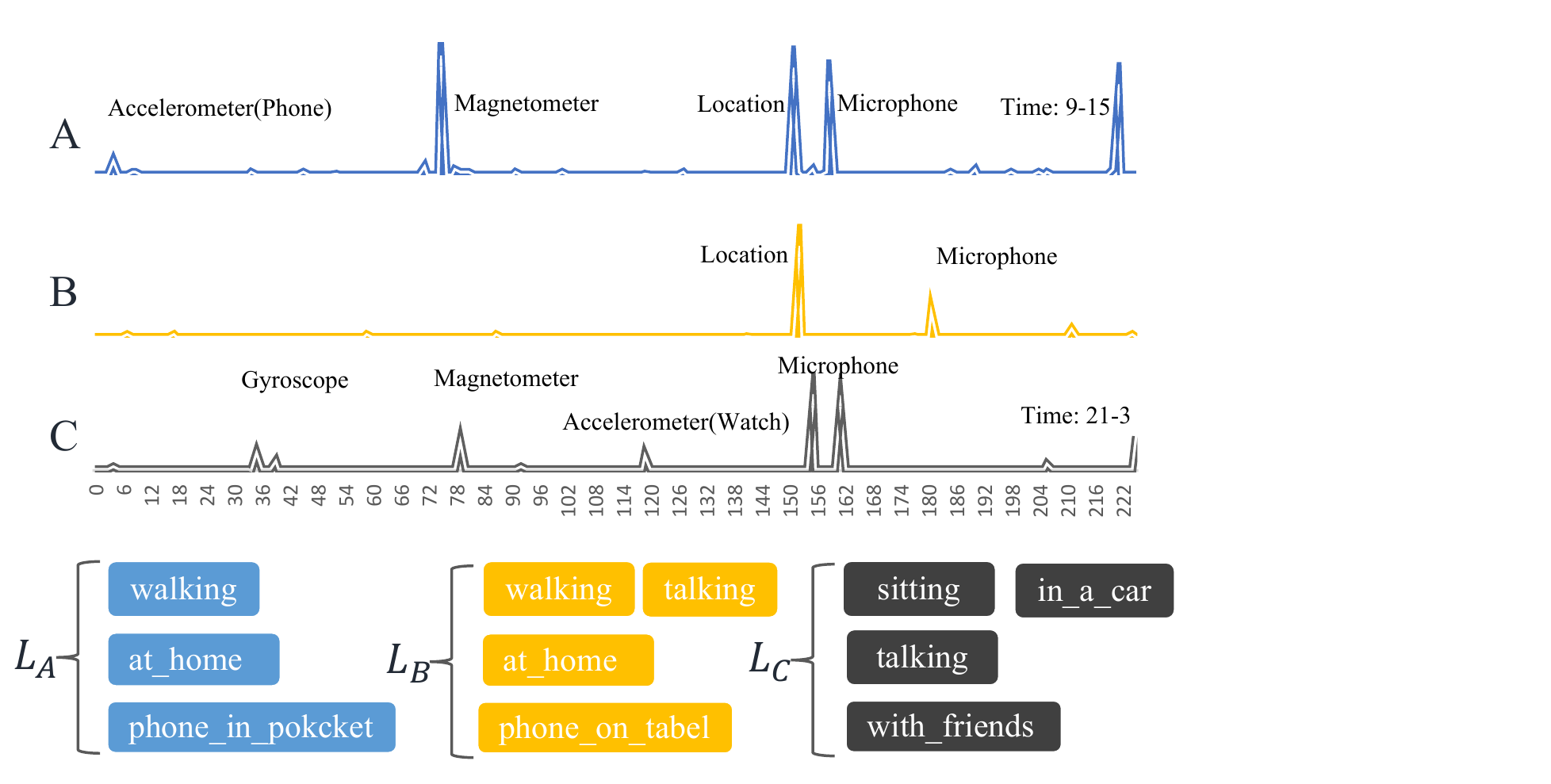}
	\caption{Case study of attention weight spikes in feature dimension captured with Sensor-specific attention. A, B, C represent three different instances at a certain timestamp. $L_A$, $L_B$, $L_C$ represent their according labels. }\label{fig:spikes}
	\end{center}
	\end{figure}
	
\subsubsection{Single-task versus Multi-task learning}
We seek to assess the benefits of multi-task learning in comparison 
to single-task learning models (See Table \ref{tab:com-result}).  
The single-task LR model has the worst F1 and Balance Accuracy score (and a  high error for  the regression model).
The single task
S-LSTM model that captures temporal dependencies 
outperforms the  LR 
model.   However, the performance of jointly 
trained multiple task learning approaches 
with LSTM (M-LSTM) is worse in comparison to the S-LSTM model. 
In general, multi-task learning approaches should improve 
classification/regression performance but fail when the 
relationships between multiple tasks are not modeled well. \model, on the other hand, outperforms the single task learning models 
because it is able to identify specific features  that 
are important (and not noisy) across the different tasks and 
across the temporal domain. 

           
\subsection{Feature Extraction}

To better understand the attention mechanisms and their ability 
to weigh certain features across the different tasks, we present 
several qualitative studies. 
%
Figure~\ref{fig:spikes} shows the burstiness of features  
(spikes) captured by the sensor-specific attention of 
three instances from three different users (tasks) at three different 
time points. We  observe 
a high correlation between 
the feature spikes
and the corresponding labels. For example, for person 
A who is walking at home with a phone in pocket;
the captured related sensors are phone 
accelerometers, magnetometer, location, microphone, and time. For person
B, who is walking but talking with a phone on the 
table, there is no change of the 
magnetometer sensor, no acceleration of the phone, and also a 
lower volume of voice (there is no position change of the phone and the device
cannot capture a higher voice on a table). For 
person C who is driving 
a car and talking with friends, the sensor-specific 
attention can capture the correlated sensors to the corresponding group activities.

	\begin{figure}[h]
 	\begin{center}
	\includegraphics[trim={2.6cm 4cm 3.7cm 2.5cm},clip, width = 8.5cm]{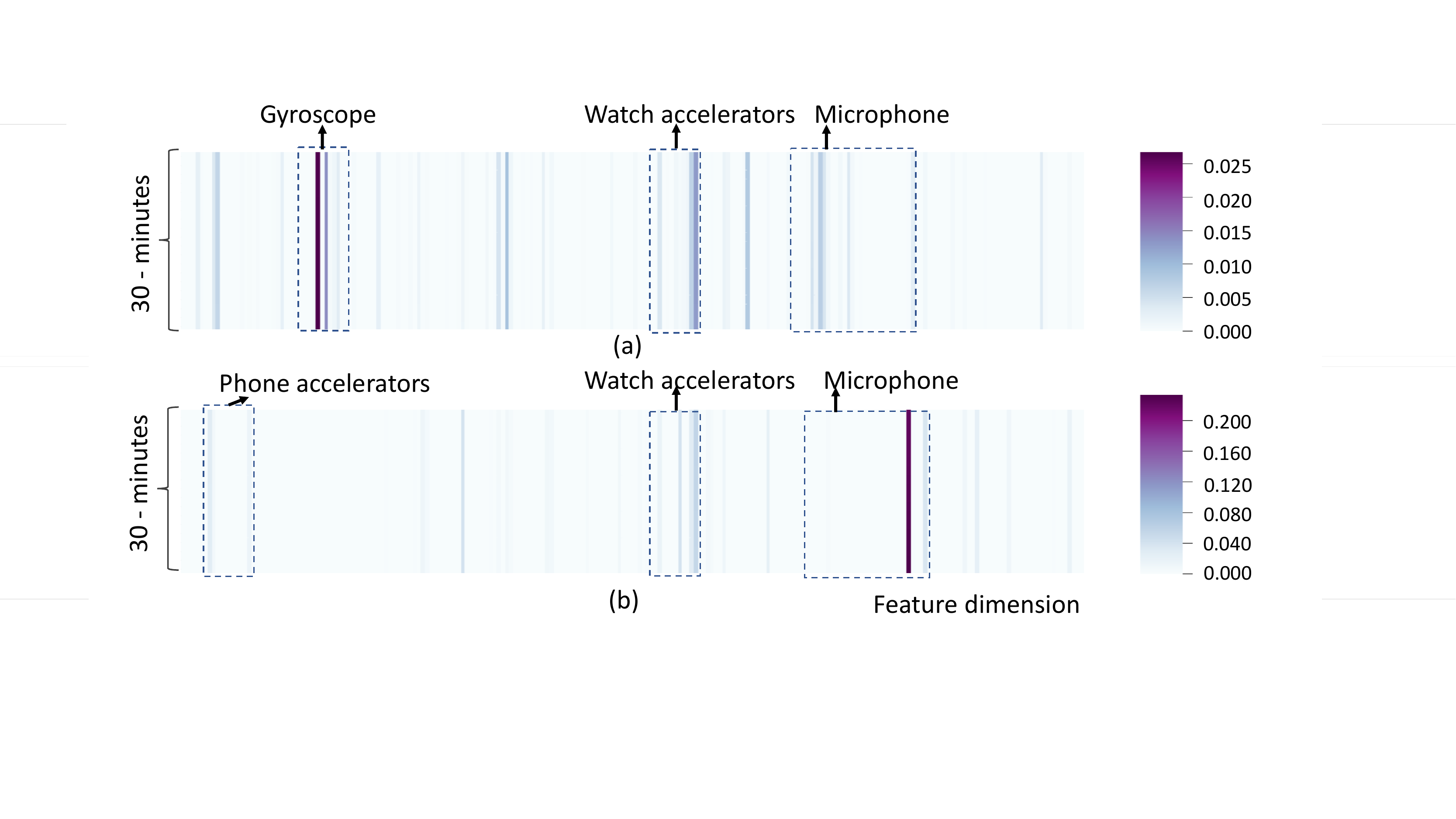}
	\caption{Sensor-specific attention matrix of two users (tasks) from ExtraSensory Dataset. Each column is the attention vector within 30-minutes time length over the input series.}\label{fig:mobile}
	\end{center}
	\end{figure}

\begin{figure}[h]
 	\begin{center}
	\includegraphics[trim={4cm 2cm 5cm 3cm},clip, width = 8.5cm]{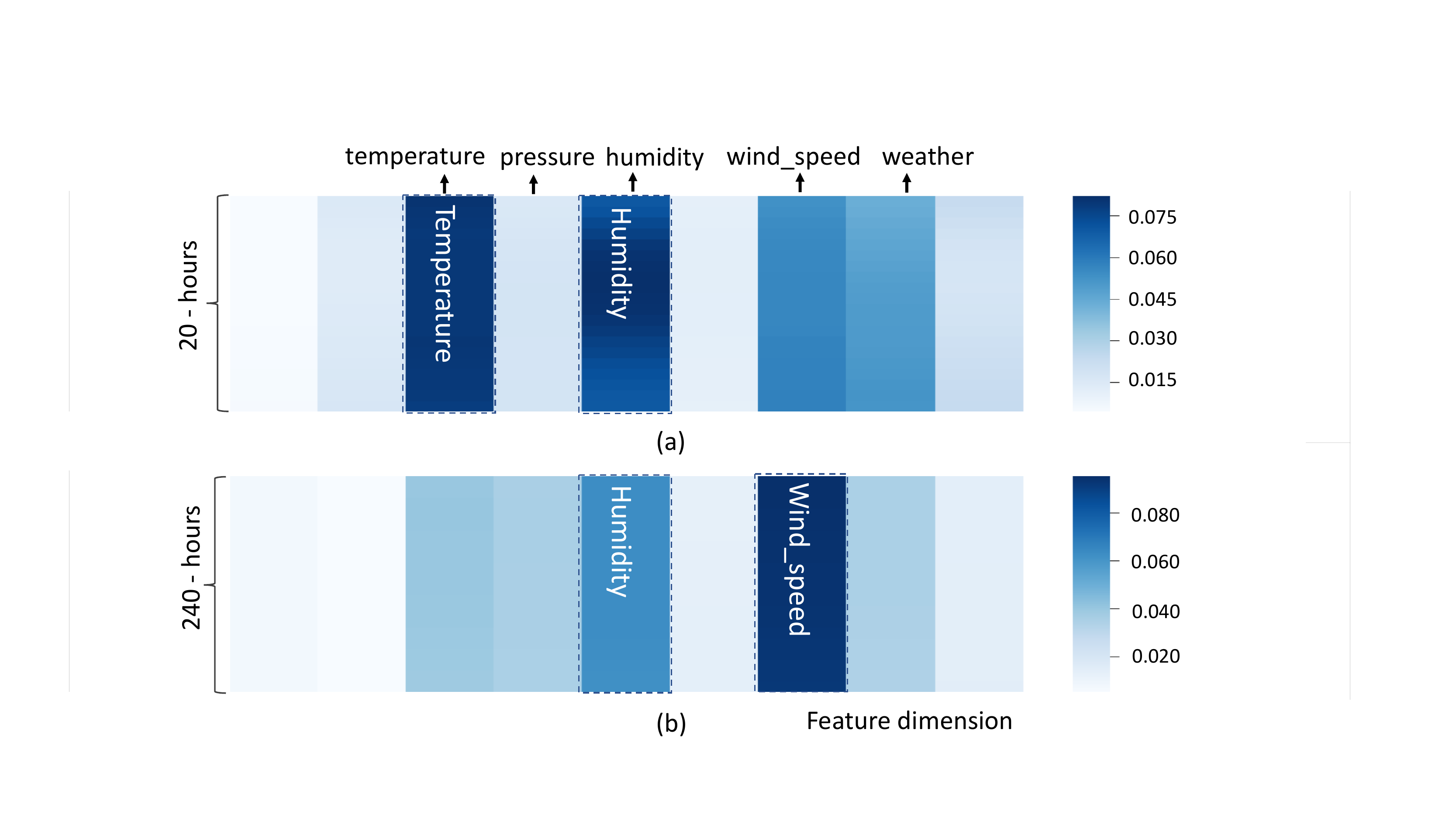}
	\caption{Sensor-specific attention matrix of one task from Air Quality Data. (a) is prediction of 20-hours time length, (b) is prediction of 240-hours time length. Each column is the attention vector over the input series.}\label{fig:weather}
	\end{center}
	\end{figure}

We take two tasks from each dataset to visualize the variation of attention vectors across feature dimensions in Figure~\ref{fig:mobile} and~\ref{fig:weather} respectively. Figure~\ref{fig:mobile} represents the attention weight matrix of two different users. 
The user of matrix (a) first lies down, then walks and talks with friends on a phone in pocket. The captured highly related sensors are phone gyroscope, watch accelerator, and microphone. 
The user of matrix (b) first grooms and gets dressed, then stays in a lab. We find that watch and phone accelerators have a strong correlation 
with body movement. The 
microphone is  directly correlated 
with voice in the surroundings.

The continuous value predictions for 
the Air Quality 
Data include 
air pollutants such as 
PM2.5, PM10, CO, NO2, O3, and SO2. From the attention weight
matrix shown in Figure~\ref{fig:weather} we observe 
that in a prediction window of 20 hours length, temperature and 
humidity have the highest weights 
amongst all the 
input features. Recall that the attention weights semantically indicate the relative 
importance of each local contributing feature series. We find that 
in case of 
short term prediction, temperature and humidity affect the air pollutants most, wind speed and weather have relatively lighter influence. This is because people in Beijing consume fuel for heating in the winter and the
humidity is usually very low during
winter time. While in a prediction window length of 240 hours, wind speed has the 
highest weight. In a dry season with low temperatures, the only effective way to disperse 
the air pollutants is wind. 
 All 
the above case studies show that our method is effective at capturing sensor-specific features and leads to interpretable results.

\subsection{Central Temporal Attention Evaluation}

\begin{figure}[h]
 	\begin{center}
	\includegraphics[trim={2cm 3cm 1.8cm 3cm},clip, width = 8.5cm]{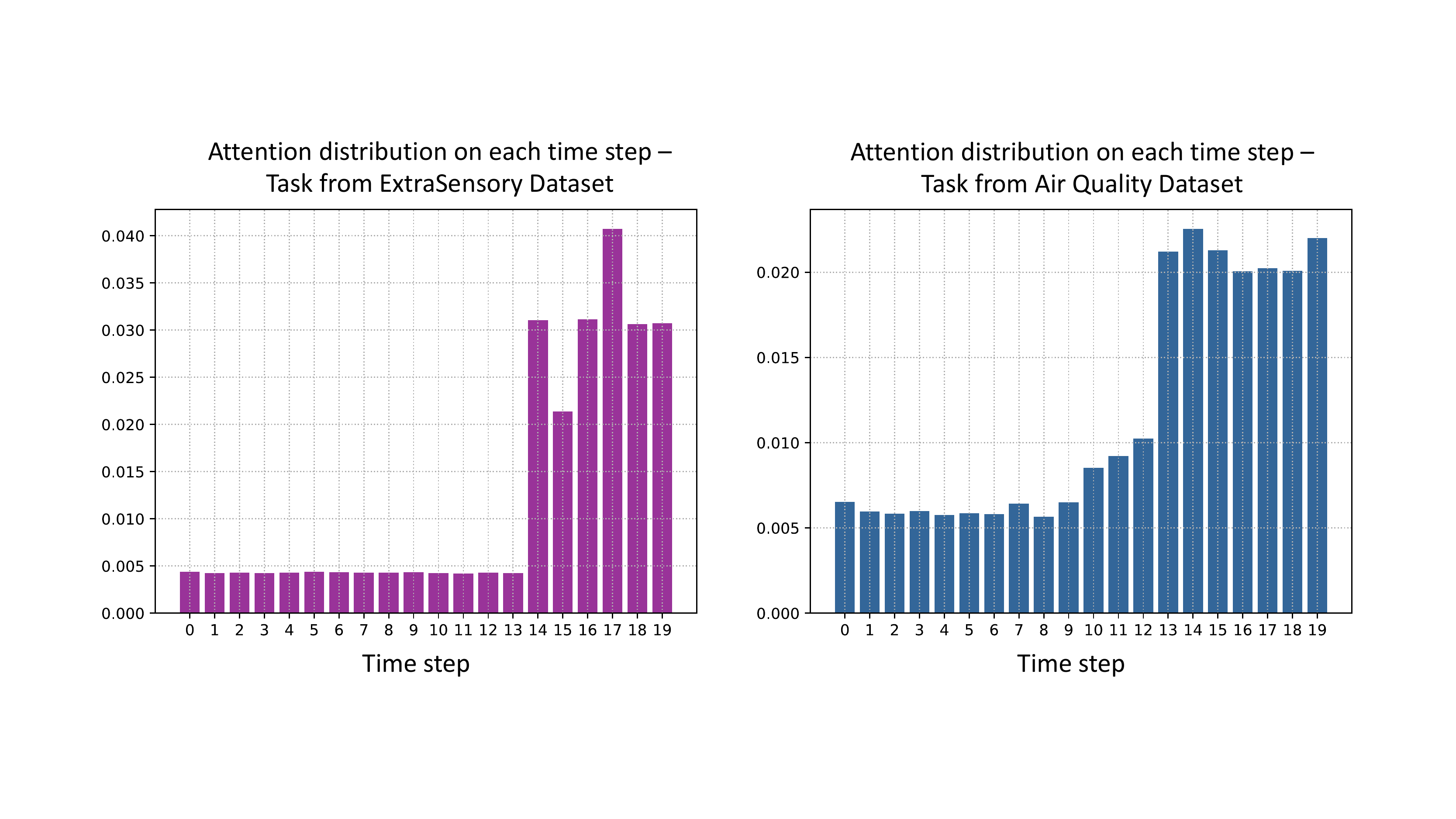}
	\caption{Time-dimension attention distribution of two different tasks}\label{fig:time-dim}
	\end{center}
	\end{figure}

The prediction performance results show the benefit of the
central time attention mechanism  (In Table \ref{tab:com-attentions}, with just central time 
  attention, \model-sa still can outperform the other baseline models).
%
  
 Figure~\ref{fig:time-dim} shows the attention distribution
  from the central time attention layer for the Extrasensory and  Air Quality datasets. 
By applying the central time attention, the weight distribution does not just focus 
  on the last step, but also  spreads to former steps. Our observation is that temporal information
  is not lost and gets re-introduced leading to a stronger predictive performance. 


\section{Conclusion}

In this paper we present \model, a novel federated 
multi-task model utilizing hierarchical attention
to generate a more efficient task-specific feature representation. Sensor-specific attention captures inner-feature correlations within 
each local task and central time attention generalizes inter-task feature representations across all tasks. We evaluate our 
proposed model on both classification and regression tasks. The 
results show that our approach achieves much better performance compared to a wide range of state-of-the-art methods. We also
show multiple qualitative 
case studies to interpret our model.
 
	In the future, we plan to investigate other federated multi-task settings such as learning multiple tasks asynchronously and mechanisms that protect local data privacy.  

\clearpage
\bibliographystyle{named}
\bibliography{ijcai19}

\end{document}